%% file: arxiv.tex
\documentclass[10pt]{article} 
\usepackage[preprint]{tmlr}

\input{math_commands.tex}

\usepackage{graphicx}
\usepackage{booktabs}
\usepackage{multirow}
\usepackage{algorithm}
\usepackage{algpseudocode}

\usepackage{url}
\usepackage{hyperref}
\usepackage{cleveref}

\title{Auto-Regressive Models Need Structural Registers: Semantic Foresight Ensures Coherent Driving Video Continuation}

\author{\name Ruibo Ming \email mrb22@mails.tsinghua.edu.cn \\
      \addr Tsinghua University, StepFun
      \vspace{-0.6em}
      \AND
      \name Jingwei Wu \email wujingwei22@mails.ucas.ac.cn\\
      StepFun
      \vspace{-0.6em}
      \AND
      \name Zhewei Huang \email hzwer@pku.edu.cn\\
      \addr StepFun
      \vspace{-0.6em}
      \AND
      \name Zhuoxuan Ju \email nymphea@stu.pku.edu.cn\\
      Peking University, StepFun
      \vspace{-0.6em}
      \AND      
      \name Daxin Jiang \email djiang@stepfun.com \\
      \addr StepFun
      \vspace{-0.6em}
      \AND
      \name Jianming Hu \email hujm@mail.tsinghua.edu.cn \\
      \addr Tsinghua University
      \vspace{-0.6em}
      \AND
      \name Lihui Peng$^{\dagger}$ \email lihuipeng@mail.tsinghua.edu.cn\\
      \addr Tsinghua University
      \vspace{-0.6em}
      \AND
      \name Shuchang Zhou$^{\dagger}$ \email shuchang.zhou@gmail.com \\
      \addr StepFun, Megvii Technology\\
      }

\def\month{MM}  
\def\year{YYYY} 
\def\openreview{\url{https://openreview.net/forum?id=XXXX}} 

\begin{document}

\maketitle

\input{sec/0_abstract}    
\input{sec/1_intro}
\input{sec/2_related}
\input{sec/3_method}
\input{sec/4_exps}
\input{sec/5_conclusion}

\bibliography{main}
\bibliographystyle{tmlr}

\appendix
\section*{Appendix}
\input{sec/X_suppl}

\end{document}

%% file: math_commands.tex
\usepackage{amsmath,amsfonts,bm}

\def\eqref#1{equation~\ref{#1}}

\def\1{\bm{1}}

\DeclareMathAlphabet{\mathsfit}{\encodingdefault}{\sfdefault}{m}{sl}
\SetMathAlphabet{\mathsfit}{bold}{\encodingdefault}{\sfdefault}{bx}{n}



%% file: sec/0_abstract.tex
\begin{abstract}
Front-view driving video continuation is a critical component for constructing sophisticated world models. However, maintaining long-term coherence faces the fundamental challenge of mitigating ``generative degeneration'' during the auto-regressive process. This phenomenon arises when the limited attention capacity of world models becomes saturated with high-entropy pixel details over extended sequences, leading to a collapse of global scene structure and motion intensity. To address this, we introduce STRIDE (STructural RegIsters for Decoupled Extrapolation), a large vision model architecture with an enforced semantic foresight mechanism. STRIDE decouples video continuation into two sub-tasks via interleaved generation of semantic and RGB tokens: forecasting low-entropy semantic maps, and subsequently generating high-frequency visual appearance conditioned on this structure. The semantic tokens explicitly function as “structural registers” -- dedicated memory slots that robustly retain the scene's long-term context and world dynamics. Furthermore, we employ an architecturally decoupled flow matching-based super-resolution module to enhance the final visual fidelity. Extensive experiments in autonomous driving scenarios demonstrate that STRIDE effectively mitigates generative degeneration, achieving minute-level, temporally consistent driving video continuation with strong generalization ability.
\end{abstract}

%% file: sec/1_intro.tex
\section{Introduction}

\begin{figure*}[t]
    \centering
    \includegraphics[width=\linewidth]{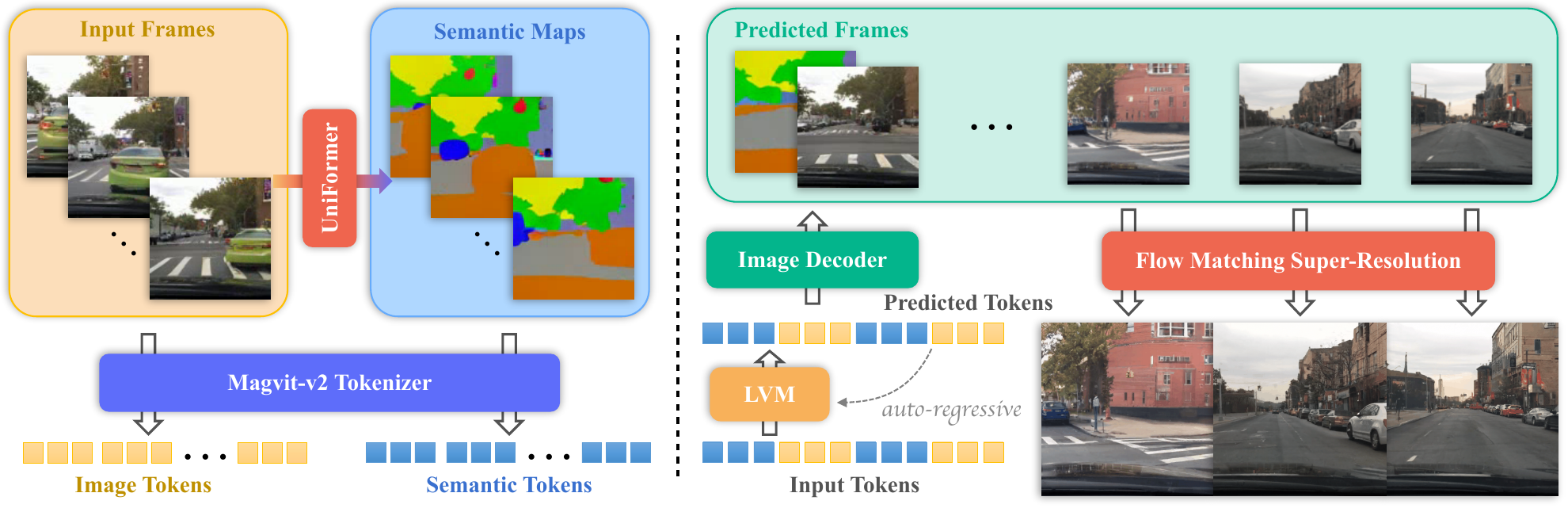}
    \caption{Overview of STRIDE architecture. The pipeline begins by processing input video frames to create two parallel data streams: the original RGB images and their corresponding semantic maps, which are estimated via a UniFormer model~\citep{li2023uniformer}. Both streams are then converted into discrete visual tokens using a unified MAGVIT-v2 tokenizer~\citep{yu2023language}. The core of the framework is a large vision model that performs auto-regressive continuation by alternately predicting future semantic and RGB tokens, where the semantic tokens serve as structural registers. Finally, the generated token sequences are decoded into video frames, and a flow matching-based super-resolution module enhances their visual quality in a dedicated post-processing stage.}
    \label{fig:pipeline}
\end{figure*}

The remarkable success of auto-regressive large language models (LLMs) in high-quality text generation has established the efficacy of the “next-token prediction” paradigm~\citep{vaswani2017attention, Radford2018ImprovingLU, brown2020language, ouyang2022training, li2025minimax, team2025kimi, guo2025deepseek}. Consequently, a significant research effort has emerged to generalize this paradigm to a broader spectrum of data modalities, utilizing discrete tokenization as the core strategy~\citep{yu2023MAGVIT, yu2023language, tian2024visual, wu2025janus, chen2025janus, ma2025janusflow, yu2026autoregressive, peng2026unified, huang2025step}. Video Continuation~\citep{oprea2020review, ming2024survey}, the task of predicting future video frames, is fundamental to building complex world models and represents a critical application of the auto-regressive approach. While modern video generation is often dominated by high-fidelity diffusion models~\citep{luo2023videofusion, wu2023tune, xing2023survey, valevski2024diffusion, cui2026selfforcing, li2026stable, lin2026mmphysvideo, xiang2026videoweave}, auto-regressive methods have historically shown greater promise for long-term motion prediction~\citep{wu2021motionrnn}. Recent breakthroughs demonstrate that purely token-based auto-regressive generation can yield effective scaling and foster non-trivial predictive capabilities in large vision models (LVMs)~\citep{sun2023emu, sun2024generative, wang2024emu3, cui2025emu3, team2025longcat, ji2026videoar}.

Pioneering works have established the potential of LVMs for video continuation~\citep{bai2024sequential, ge2024worldgpt, kondratyuk2023videopoet}, yet the prohibitive computational demands and inherent constraints in handling long sequences severely limit thorough exploration. Specifically, front-view driving video continuation demands that models maintain long-term temporal coherence for global scene structure and object dynamics~\citep{hu2023gaia, russell2025gaia, wang2026fine, xie2026raynova, zhang2026horizondrive, mittal2026orbis}. However, maintaining this coherence faces the fundamental challenge of generative degeneration during the auto-regressive process~\citep{Holtzman2020The, welleck2019neural, han2026entropy, zou2026hiar, huang2026live, li2026freqforcing, cui2026lol, chen2026mitigating}. This degeneration phenomenon occurs because high-frequency visual information consumes a disproportionate amount of attention capacity, preventing the model from effectively focusing on the low-entropy, long-range structural information. This imbalance leads to a collapse of the global scene structure and motion intensity, plunging the generation into stationary or repetitive outputs.

To address the generative degeneration problem, we introduce STRIDE, a novel video continuation framework that enforces a semantic foresight mechanism through structural registers. STRIDE is built upon a large auto-regressive transformer trained on large-scale tokenized front-view driving videos (illustrated in \figurename~\ref{fig:pipeline}). STRIDE uniquely decouples the complex video continuation task into two more constrained sub-tasks via interleaved generation of semantic and RGB tokens:

\begin{enumerate}
    \item Forecasting the low-entropy structural blueprint (semantic map). 
    \item Subsequently generating the high-frequency visual appearance conditioned on this predicted structure.
\end{enumerate}

\begin{figure*}[t]
    \centering
    \includegraphics[width=\linewidth]{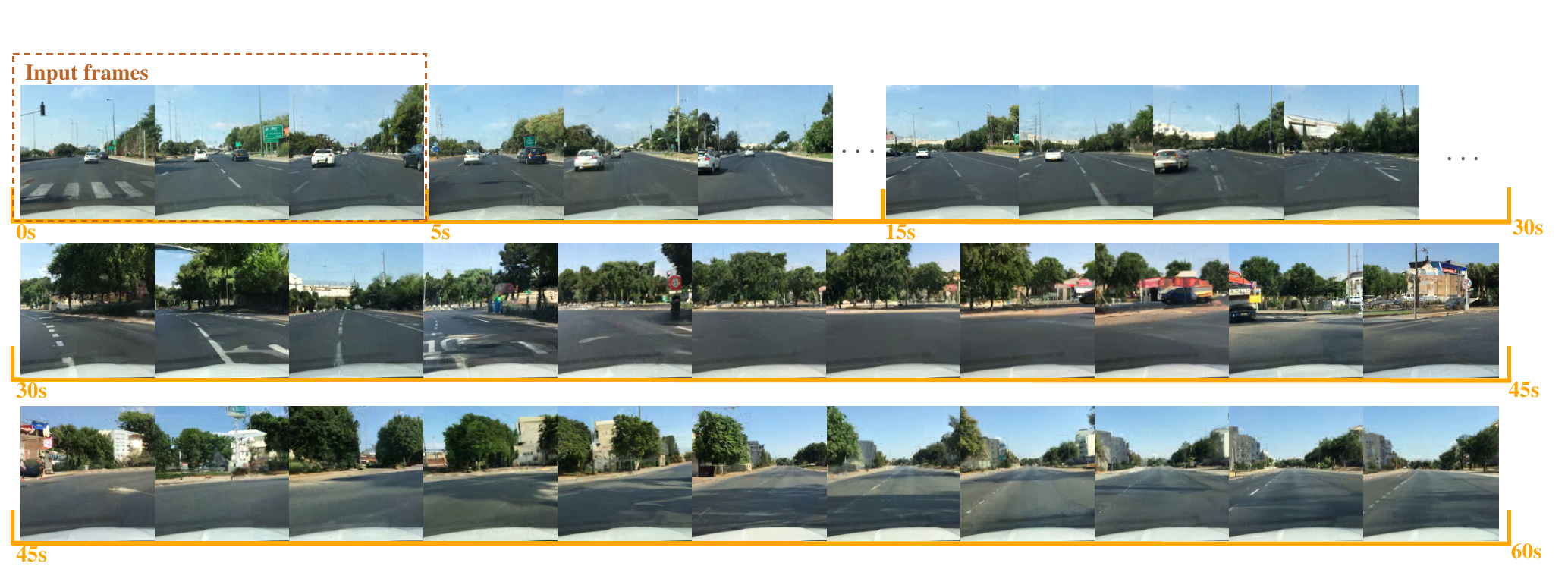}
    \caption{Coherent generation of a minute-level driving maneuver. This figure showcases STRIDE generating a continuous sequence of $45$ frames from just $3$ input frames (sampled at $0.6$ Hz). The generated video depicts a complex and logical driving maneuver where the ego-vehicle proceeds forward, changes lanes, and executes a right turn. STRIDE coherently synthesizes new environmental elements throughout the entire process.}
    \label{fig:auto}
\end{figure*}

We formally posit that the semantic tokens explicitly function as structural registers. This concept stems from the observation that Vision Transformers (ViTs) often repurpose redundant tokens as implicit registers for global information processing~\citep{darcet2024vision, yan2025vision, jiang2025vision, shi2026vision, starodubcev2026registers, hong2026test, kirby2026driving}. In STRIDE, the semantic tokens serve as dedicated memory slots that robustly retain the scene's long-term structural context and world dynamics. STRIDE streams the video's structural information as a token sequence through the generative model, requiring the model to first predict the semantic structure of the frame before predicting its corresponding visual content (RGB) at each timestep. This enforced semantic foresight mechanism, via dual-modality interleaved generation, effectively mitigates the generative degeneration phenomenon in auto-regressive long-term video generation. This architectural guidance mimics cognitive planning~\citep{hayes1979cognitive}, freeing the LVM from high-entropy pixel complexity and enabling it to focus its attention capacity on maintaining long-range coherence. Visual fidelity is further enhanced by a flow matching-based super-resolution module, implemented as a dedicated post-processing step to decouple the burden of generating visual details from the LVM. As a result, our framework generates coherent, minute-level front-view driving videos from a brief initial sequence, shown in \figurename~\ref{fig:auto}. Experiments validate the effectiveness of our framework, demonstrating impressive quantitative and qualitative results.

We introduce the STRIDE framework, addressing the core limitations of LVMs in long-term video continuation. Our main contributions are summarized as follows:

\begin{enumerate}
    \item Structural registers for long-term coherence: We define interleaved semantic tokens as structural registers, offering a dedicated memory mechanism that enhances temporal consistency and facilitates the learning of complex world dynamics.
    \item Mitigating generative degeneration via semantic foresight: We introduce the STRIDE framework, which implements a semantic foresight mechanism through interleaved dual-modality tokens, fundamentally mitigating generative degeneration in long-term front-view driving video continuation.
    \item Architecturally decoupled high visual fidelity and computational efficiency: We validate an architecturally decoupled design using a post-processing, flow matching-based super-resolution module. This module plays a crucial role in enhancing visual fidelity while significantly alleviating the computational burden on the LVM. 
\end{enumerate}

%% file: sec/2_related.tex
\section{Related work}

\paragraph{Discrete visual tokenizers.} A foundational paradigm for applying auto-regressive models to visual generation is the vector quantization of continuous images into discrete tokens~\citep{van2017neural, razavi2019generating, esser2021taming, zheng2022movq, chang2022maskgit, mentzer2024finite, zhao2025image, zheng2026taming}. This research area features diverse tokenization strategies that optimize for different objectives: one-dimensional tokenizers prioritize a low token rate to maximize compression~\citep{yu2024image, kim2025democratizing, atanov2026videoflextok}, while unified tokenizers focus on learning a unified representation for multimodal alignment and joint training~\citep{mizrahi20244m, bachmann20244m, ma2025unitok, wang2024omnitokenizer, peng2026unified, yao2026unified}. Nevertheless, the reconstruction quality based on discrete tokenizers is generally inferior to continuous latent diffusion-based models due to the introduction of quantization errors. HART employs a diffusion model to construct a hybrid architecture for recovering details lost during tokenization~\citep{tang2024hart}. MAGVIT mitigates the fidelity reduction inherent in such quantization by adopting a lookup-free quantization method, which enables a super-large codebook~\citep{yu2023MAGVIT, yu2023language, luo2024open}. Consequently, the tokenizer exhibits exceptionally high reconstruction fidelity even under strong spatial compression. Moreover, its temporally causal convolutional architecture and vector-quantized representation make MAGVIT particularly suitable for auto-regressive visual generation frameworks~\citep{kondratyuk2023videopoet, wang2025bridging, seawead2025seaweed}. VAR~\citep{tian2024visual, han2025infinity, ji2026videoar} enhances reconstruction quality by switching to a next-scale prediction mode, iteratively predicting tokens from low- to high-resolution. GigaTok~\citep{xiong2025gigatok} demonstrates that simply scaling the tokenizer can also significantly improve reconstruction. LlamaGen~\citep{sun2024autoregressive} and VA-VAE~\citep{yao2025reconstruction} both reveal a fundamental trade-off between reconstruction fidelity and generation quality in visual tokenizers, underscoring the importance of tokenizer architectures that can effectively balance both objectives.

\paragraph{Auto-regressive visual generation.} The exploration of the next-token prediction paradigm for visual data is driven by its inherent advantages, such as architectural scalability and compatibility with multi-modal fusion~\citep{yan2021videogpt, villegas2022phenaki, gupta2023photorealistic, liu2024world, wang2024loong, polyak2024movie, yu2026mirai}. Early seminal work like DALL-E has demonstrated the viability of this approach for text-to-image generation by tokenizing images with a dVAE~\citep{ramesh2021zero}. Sequential Large Vision Models demonstrate that auto-regressive learning on purely tokenized visual data can yield effective scaling and non-trivial predictive capabilities~\citep{bai2024sequential}. WorldGPT~\citep{ge2024worldgpt} shows the capacity to acquire an understanding of world dynamics from large-scale video analysis. This line of research establishes auto-regressive models as a powerful architecture for simulating future environments. Advanced systems achieve industry-level results in long-term, token-based video generation~\citep{kondratyuk2023videopoet, hu2023gaia, russell2025gaia, team2025longcat}.

\paragraph{Multimodal in-context learning.} A prominent trend in auto-regressive modeling involves unifying diverse data modalities (e.g., RGB image, segmentation map, depth map, optical flow, sketch, text, audio) into a shared, discrete token space~\citep{girdhar2023imagebind, zheng2024unicode}. Such a unified space has enabled powerful in-context learning applications, including cross-modal translation~\citep{kondratyuk2023videopoet, bai2024sequential, mizrahi20244m, bachmann20244m} and, more relevant to our work, the use of auxiliary modalities as control signals for video generation~\citep{wang2023images, wang2023seggpt, team2024chameleon, wang2022internvideo, wang2024internvideo2, yao2024car, yang2026geniedrive, Rahimi_2026_CVPR}. The effectiveness of auxiliary modalities stems from their ability to represent high-level structural information at a much lower information density than raw pixels.

%% file: sec/3_method.tex
\section{Method}

The task of video continuation is defined as generating a sequence of future frames $\{\tilde{I}_{t+1}, \tilde{I}_{t+2}, \tilde{I}_{t+3}, ...\}$ conditioned on $t$ past frames $\{I_{i}\in \mathbb{R}^{h\times w\times 3}|_{i=1,...,t}\}$. Our model is prompted with three consecutive frames and operates within an auto-regressive paradigm, enabling the iterative generation of an arbitrary number of subsequent frames. This capability allows STRIDE to produce creative, minute-level videos. As illustrated in \figurename~\ref{fig:pipeline}, our framework consists of three main components: (1) a visual tokenizer that encodes both images and semantic maps into discrete tokens; (2) a large vision model that performs the video continuation task by alternately predicting semantic and RGB tokens; (3) a flow matching-based super-resolution module that enhances the visual fidelity of decoded frames.

\subsection{Visual tokenizer}

Our framework necessitates a visual tokenizer that effectively balances two competing needs: the high reconstruction fidelity necessary for visual quality and a high spatial compression rate to make sequence modeling computationally tractable for large vision models~\citep{yu2023language, sun2024autoregressive}. To address this, we adopt the MAGVIT-v2 architecture~\citep{yu2023language}, utilizing the publicly available weights from Open-MAGVIT2~\citep{luo2024open}. Its lookup-free quantization method enables the use of an extremely large vocabulary (size of $2^{18}$). This massive vocabulary is critical, as it enables a high spatial compression rate (e.g., an $8 \times$ down-sampling) while simultaneously retaining the exceptional reconstruction fidelity essential for high-quality video generation.

\subsection{Large vision model}

We adopt the LLaMA architecture~\citep{touvron2023llama} for our large vision model. To make the large vocabulary of MAGVIT-v2 tokenizer computationally tractable, we leverage its token factorization technique~\citep{yu2023language, luo2024open}. This method decomposes the whole $2^{18}$-sized codebook into two smaller, concatenated $2^{9}$-sized codebooks, significantly reducing computational complexity while retaining expressive capacity. The visual tokenizer converts each $112 \times 112 \times 3$ image (either an RGB frame or a semantic map) into a sequence of $392$ tokens. For semantic foresight training, the token sequences are interleaved to create a $784$-token representation per semantic-RGB pair. The LVM is trained from scratch with a context length of $16{,}384$ tokens, accommodating sequences of up to $20$ such pairs. Its training objective is the standard next-token prediction task, optimized by minimizing the cross-entropy loss:

\begin{equation}
    \mathcal{L}_{\text{LVM}} = -\sum_{i=1}^{N} \log P(z_i | z_{<i}; \theta)
    \label{eq: lvm}
\end{equation}

\subsection{Structural registers}

Our STRIDE framework is built upon the core principle of interleaved generation, which we frame as an enforced semantic foresight mechanism with structural registers. Instead of directly predicting the next RGB frame from a history of past RGB frames, we decompose the task by introducing an intermediate representation: semantic maps. This approach hypothesizes that predicting high-level structure (semantics) first, and then translating the visual appearance conditioned on that structure, is a more tractable learning problem for long-term video continuation.

Formally, let $I_t$ and $S_t$ denote the RGB frame and its corresponding semantic map at timestep $t$. A visual tokenizer $\mathcal{T}$ converts them into discrete token sequences, $Z^I_t = \mathcal{T}(I_t)$ and $Z^S_t = \mathcal{T}(S_t)$. A standard auto-regressive model would predict the next frame by modeling the probability $P(Z^I_t | Z^I_{<t})$. The interleaved generation process, however, models the joint probability. Let $H_{<t} = (Z^I_{<t}, Z^S_{<t})$ represent the full history of tokens before timestep t, the joint probability is defined as~\cref{eq: interleaved}.

\begin{equation}
    P(Z^I_t, Z^S_t | H_{<t}) = P(Z^S_t | H_{<t}) \cdot P(Z^I_t | H_{<t}, Z^S_t)
    \label{eq: interleaved}
\end{equation}

During inference, at each timestep, the model first predicts the next semantic token sequence $\hat{Z}^S_t$, and then predicts the RGB token sequence $\hat{Z}^I_t$ conditioned on both the history $H_{<t}$ and the newly generated semantic tokens $\hat{Z}^S_t$.

This decomposition of the joint probability is hypothesized to simplify the learning problem. It effectively reframes the video continuation task into two more constrained sub-tasks: (1) semantic forecasting, where the model predicts the future semantic layout, and (2) conditional translating, where it generates the visual appearance based on that predicted structure and historical images. This approach enables the model to reason about and preserve high-level structural information explicitly. By mandating the prediction of structure before appearance, the structural registers introduce a crucial inductive bias into the generation process. As a result, it enhances temporal consistency and helps prevent generative degeneration into stationary or repetitive scenes.

\subsection{Flow matching-based super-resolution}

The post-processing super-resolution module serves two critical goals: mitigating the quality loss inherent in spatial compression tokenization and significantly improving computational efficiency. We adopt a flow matching-based approach, which learns a direct, continuous transformation from a low-resolution ($112 \times 112$) input to a high-resolution ($448 \times 448$) output by modeling a velocity vector field governed by an ordinary differential equation (ODE)~\citep{schusterbauer2024fmboost}. Since the number of tokens per image grows quadratically with image resolution, this architectural decoupling strategy saves $16 \times$ the training cost and improves $16 \times$ the inference efficiency, at the expense of only a super-resolution module whose size is negligible compared to the LVM. This decoupling ensures that the LVM focuses its capacity exclusively on the challenging task of sequence modeling and coherence, while the specialized super-resolution module handles the deterministic generation of high-frequency visual fidelity.

Given a low-resolution image, it is first upsampled via bilinear interpolation to get a blurry image $x_l$. A residual image is defined as $r = x_h - x_l$, where $x_h$ is the high-resolution target. The objective of the flow matching model is to predict $r$ by learning a continuous path $x_t$ over the time interval $t \in [0, 1]$. This path starts from a noisy image $z$ (created by adding noise to $x_l$) at $t = 0$, and ends at the target residual $r$ at $t = 1$. The evolution along this path is guided by a time-conditioned EfficientUNet $v_\theta(x_t, c, t)$~\citep{saharia2022photorealistic}, where the blurry image $x_l$ serves as the context $c$. The network is trained to approximate the optimal vector field $u_t = \frac{dx_t}{dt}$ by minimizing the flow matching loss defined as~\cref{eq: fm}.

\begin{equation}
    \mathcal{L}_{\text{SR}}=\mathbb{E}_{t\sim\mathcal{U}(0,1)}[||v_{\theta}(x_t, c,t)-u_t||_{2}^{2}]
    \label{eq: fm}
\end{equation}

During inference, a generated low-resolution frame is first upsampled via bilinear interpolation to produce the blurry base image $x_l$. We then use the 4th-order Runge-Kutta method~\citep{dormand1980family} as a numerical ODE solver to integrate the learned vector field $v_\theta$. Given the sampled initial state, this integration deterministically produces the high-frequency residual image $r$, which is then added to $x_l$ to produce the final, high-fidelity output frame $\hat{x_h}$.

%% file: sec/4_exps.tex
\section{Experiments}

\begin{table}[t]
    \centering
    \caption{Codebook usage efficiency. Comparison of the codebook percentage required by RGB versus semantic tokens to cover different proportions of the training dataset. Semantic tokens demonstrate significantly higher efficiency, representing the majority of the data with a much smaller fraction of the total codebook.}
    \begin{tabular}{@{}lrrr@{}}
        \toprule
        \multirow{2}{*}{Modality} & \multicolumn{3}{c}{Training Data Coverage} \\
        \cmidrule(l){2-4} 
         & 50\% & ~~~~~~95\% & 99\% \\
        \midrule
        RGB Codebook Usage & 23\% & ~~~~~~82\% & 94\% \\
        Semantic Codebook Usage & 1\% & ~~~~~~30\% & 54\% \\
        \bottomrule
        \end{tabular}
    \label{tab:codebook}
\end{table}

This section presents a comprehensive evaluation of the STRIDE framework on the long-term video continuation task. Quantitative and qualitative results demonstrate that the proposed semantic-guidance scheme is crucial for generating creative and temporally consistent long videos. Furthermore, we verify that the architecturally decoupled super-resolution module effectively enhances final visual quality while alleviating the burden of texture synthesis from the LVM.

\subsection{Experimental setup}

\paragraph{Datasets.} We choose BDD100K~\citep{yu2020bdd100k} as our model's training data, a large-scale dataset of first-view driving videos. It contains $100{,}000$ videos, most of which are at $30$ FPS, with video duration generally in the $30$ to $40$ seconds range. In the training stage, we only use its official training set, which contains $70{,}000$ videos totaling about $700$ hours. This dataset was selected for its significant data diversity and average video duration suitable for exploring creative, long-term generation. During data pre-processing, all videos are downsampled to $3$ Hz and frames are center-cropped to $112 \times 112$ resolution. The visual tokenizer converts each image, whether an RGB frame or a semantic map, into a sequence of $392$ tokens. The final training corpus, containing interleaved RGB and semantic sequences, totals approximately $6$ billion tokens. For evaluation, the STRIDE framework is assessed at an inference frequency of $0.6$ FPS on $2$ benchmarks: $100$ randomly selected videos from the BDD100K test set and all $150$ videos from the nuScenes validation set~\citep{caesar2020nuscenes}.

\paragraph{Training procedure.} To assess the impact of semantic guidance, we train a single model on a mixture of two configurations: (1) a baseline using only RGB tokens, and (2) our proposed STRIDE method using interleaved semantic and RGB tokens; a system prompt selects which mode the model runs at inference. Each training sample consists of a sequence of $18$ timesteps. For the RGB-only baseline, one timestep corresponds to the $392$ tokens from a single frame; for the STRIDE method, one timestep consists of $784$ tokens created by interleaving the tokens from dual modalities. The global batch size is set as $64$, and the learning rate is linearly annealed from $10^{-5}$ to $2\times 10^{-6}$. The probing model with 7B parameters is trained for $20{,}000$ iterations on $256$ A800 GPUs, totaling approximately $12$ hours. The total number of image tokens being processed in the training stage is approximately $14$ billion. The main model with 20B parameters triples the number of LVM layers and is trained for $50{,}000$ iterations with other hyperparameters unchanged, totaling approximately $90$ hours. The total number of image tokens being processed in the training stage is approximately $35$ billion.  With the 7B model, the inference speed is about $60$ tokens per second.

\subsection{Probing experiments}

We begin our experimental analysis by conducting probing experiments on the smaller, 7B variant of our STRIDE model. This approach allows us to isolate and efficiently analyze the fundamental impact of the semantic-guidance scheme on generative behavior, before scaling to the full 20B model.

\begin{figure}[t]
    \centering
    \includegraphics[width=0.6\linewidth]{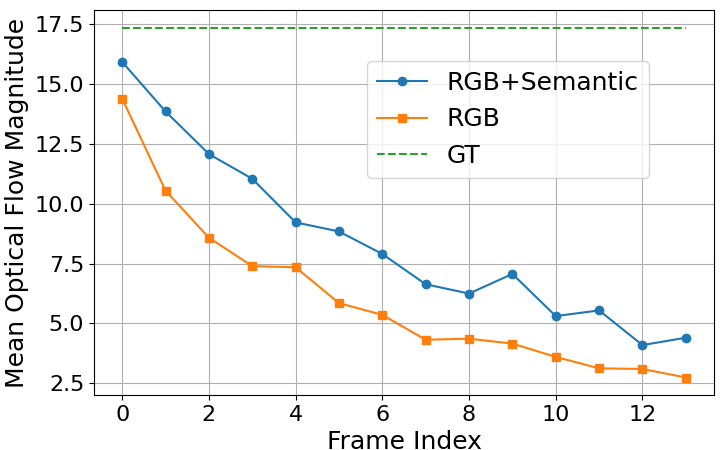}
    \caption{Semantic foresight mitigates generative degeneration. The mean optical flow magnitude is plotted for $150$ $15$-frame clips generated on the nuScenes validation set~\citep{caesar2020nuscenes}. Lower values indicate less motion, a key symptom of the degeneration phenomenon. The model with semantic guidance maintains significantly more motion than the RGB-only baseline, validating that our STRIDE scheme effectively mitigates degeneration. The ground truth (GT) from real data is shown for reference.}
    \label{fig:probing}
\end{figure}

\paragraph{Efficiency in representation.} As illustrated in \tablename~\ref{tab:codebook}, semantic tokens demonstrate a significantly higher efficiency in data representation compared to RGB tokens. An analysis of the BDD100K training dataset reveals that to cover $50\%$ of the training data, semantic tokens require only $1\%$ of the codebook, whereas RGB tokens consume $23\%$. This efficiency gap widens at higher coverage levels: encompassing $99\%$ of the semantic data requires only $54\%$ of the codebook, while RGB data requires $94\%$. These findings highlight that semantic tokens, being more abstract, provide a much more compact representation of the underlying visual information.

\paragraph{Guidance in generation.} The most significant benefit of integrating the structural registers is hypothesized to be the mitigation of generative degeneration, a common failure mode in auto-regressive models where motion progressively stalls. We propose that by providing the LVM with an explicit, low-density structural prior (i.e., semantic maps), the model can maintain long-term scene coherence and object permanence more robustly. This is because the structural registers act as a powerful constraint, forcing the model to forecast scene dynamics structurally rather than relying solely on the noisy, high-entropy pixel values. To quantitatively validate this hypothesis, we measure the structural registers' impact on the degeneration phenomenon. We use the mean magnitude of optical flow between adjacent frames, computed with a pre-trained RAFT model~\citep{teed2020raft}, as a proxy for motion intensity in generated videos. As shown in~\figurename~\ref{fig:probing}, the model with semantic guidance maintains significantly larger motion intensity over time compared to an RGB-only baseline, confirming that our STRIDE scheme effectively mitigates the generative degeneration phenomenon.

\subsection{Quantitative results}

\begin{table}[t]
  \centering
  \caption{Quantitative comparison on the nuScenes validation set. STRIDE achieves the best FVD score among all compared methods, even without being fine-tuned on the nuScenes dataset. $\dagger$ denotes general-purpose video generation models.}
  \begin{tabular}{@{}lcc@{}}
    \toprule
    Methods & FID $\downarrow$ & FVD $\downarrow$ \\
    \midrule
    \textit{nuScenes fine-tuning} \\
    DriveDreamer~\citep{wang2024drivedreamer}& 52.6 & 452 \\
    DrivingDiffusion\citep{li2024drivingdiffusion} & 15.6 & 335 \\
    SubjectDrive~\citep{huang2024subjectdrive} & 16.0 & 124 \\
    Panacea~\citep{wen2024panacea} & 17.0 & 139.0 \\
    Drive-WM~\citep{wang2024driving} & 15.8 & 122.7 \\   
    DriveDreamer-2~\citep{zhao2024drivedreamer} & 18.4 & 74.9 \\
    Vista~\citep{gao2024vista} & \textbf{6.9} & 89.4 \\
    WoVoGen~\citep{lu2025wovogen} & 27.6 & 417.7 \\
    \midrule
    \textit{w/o nuScenes fine-tuning} \\
    DriveGAN~\citep{kim2021drivegan} & 73.4 & 502.3 \\
    GenAD~\citep{yang2024generalized} & 15.4 & 184.0 \\
    DrivingWorld~\citep{hu2024drivingworld} & 16.4 & 174.4 \\
    LVM~\citep{bai2024sequential} $\dagger$ & 71.5 & 162.2 \\
    StreamingT2V~\citep{henschel2024streamingt2v} $\dagger$ & 28.8 & 131.0 \\
    CogVideoX~\citep{yang2024cogvideox} $\dagger$ & 64.9 & 79.8 \\
    STRIDE & 16.7 & \textbf{40.8} \\
    \bottomrule
  \end{tabular}
  \label{tab:main}
\end{table}

We evaluate the STRIDE model's performance on the video continuation task against related video generation methods. As shown in~\tablename~\ref{tab:main}, STRIDE achieves a state-of-the-art Fréchet Video Distance (FVD) score~\citep{unterthiner2018towards} on the nuScenes validation set. This result is achieved without any training or fine-tuning on the nuScenes dataset, demonstrating our model's strong generalization ability in addition to its capacity for generating high-quality videos.

\subsection{Qualitative results}

\begin{figure*}[t]
    \centering
    \includegraphics[width=1.0\linewidth]{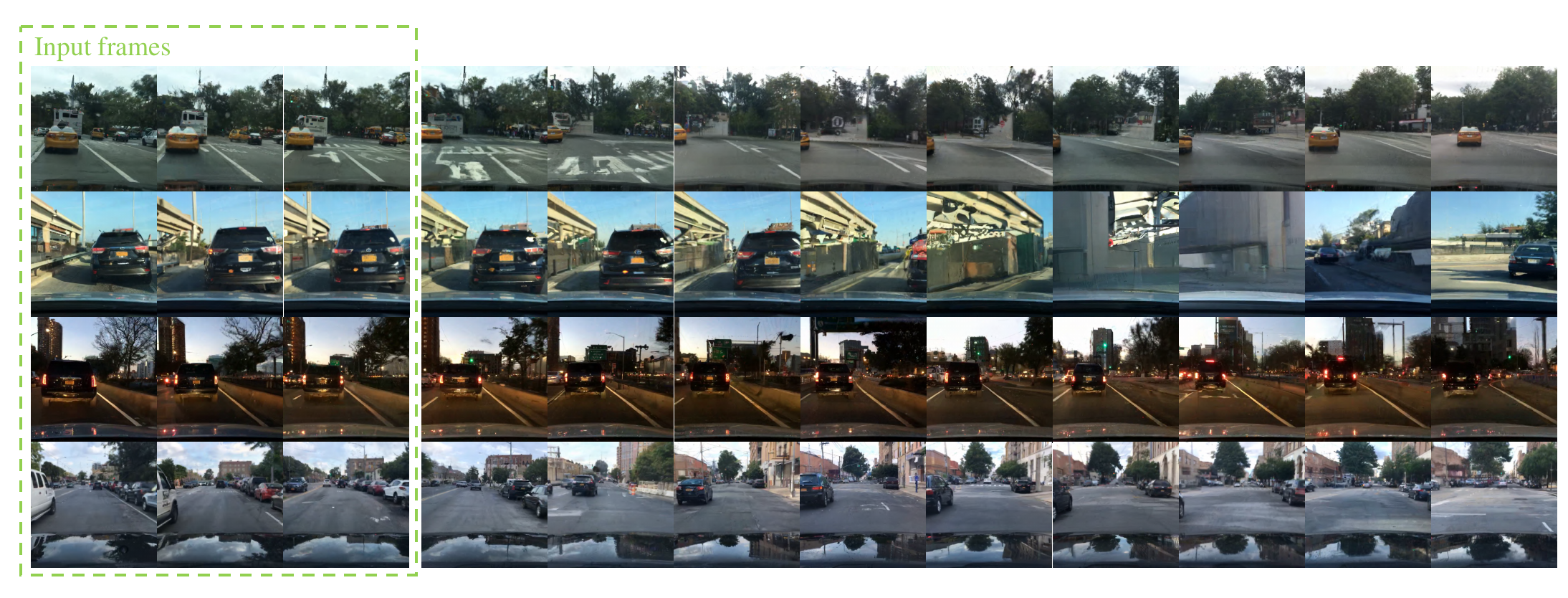}
    \caption{Qualitative analysis of generated videos. Four diverse scenarios generated by STRIDE are presented. From top to bottom: (1) The model correctly infers the ego-vehicle's turn by following the lead vehicle, demonstrating object permanence after the turn. (2) An edge case is generated where the ego-vehicle executes a U-turn in response to a stopped lead vehicle. (3) The model demonstrates an understanding of dynamic object interactions by accurately activating the brake lights of a lead vehicle as it decelerates. (4) Novel scenes are created with physically plausible details, including consistent hood reflections.}
    \label{fig:vis}
\end{figure*}

\begin{figure}[h!]
    \centering
    \includegraphics[width=0.6\linewidth]{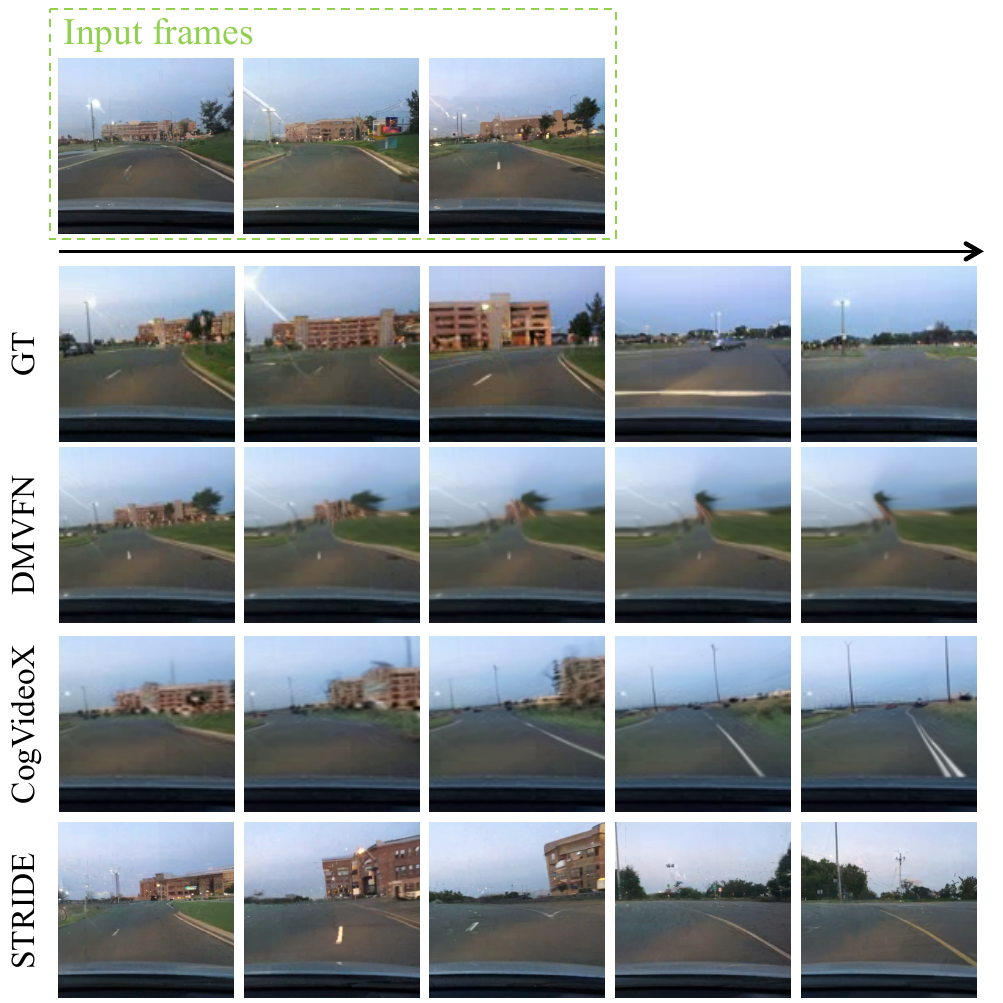}
    \caption{Qualitative comparison against related methods. Compared to the optical flow-based method DMVFN~\citep{hu2023dynamic}, which struggles to generate novel scenes, and the diffusion transformer-based model CogVideoX~\citep{yang2024cogvideox}, which is prone to motion blur, STRIDE better preserves object structure while generating coherent new scenes.}
    \label{fig:compare}
\end{figure}

Unless otherwise specified, all qualitative results presented are generated by conditioning the model on the first $3$ frames of a given sequence, and sampled from the BDD100K dataset at $0.6$ Hz.

\begin{figure}[h!]
    \centering
    \includegraphics[width=0.6\linewidth]{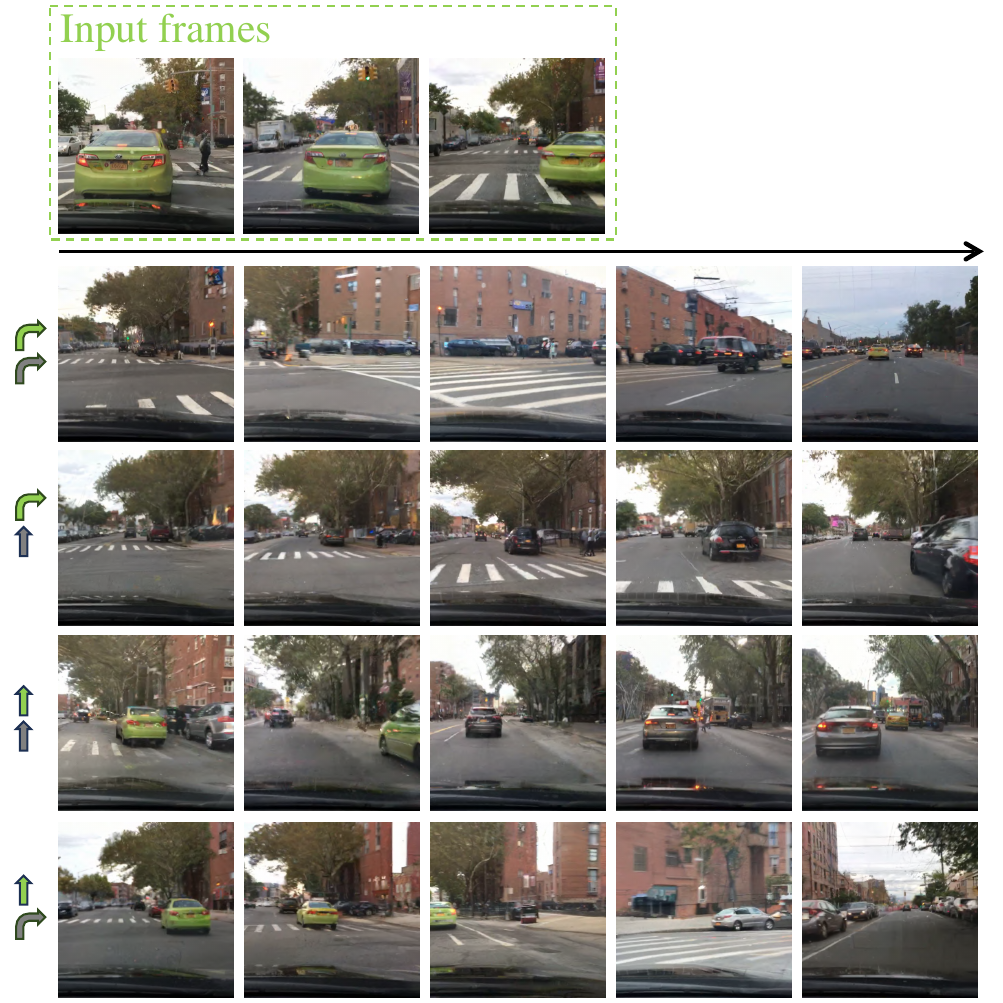}
    \caption{Stochastic generation of diverse future timelines. From an identical $3$-frame prompt, the model stochastically generates multiple plausible scenarios. The leftmost arrows indicate the driving directions of the lead vehicle (top) and the ego car (bottom) for each generated timeline.}
    \label{fig:timelines}
\end{figure}

\paragraph{Coherent and creative video continuation.} STRIDE demonstrates a strong capacity for video continuation from a moving vehicle's perspective, generating scenarios that are both physically plausible and logically consistent. A notable emergent capability is the model's implicit acquisition of fundamental traffic knowledge. For instance, it respects lane discipline by rarely attempting incorrect turns (e.g., a left turn from a right-turn lane) and exhibits collision-avoidance behavior by decelerating or maneuvering around lead vehicles. STRIDE also excels at creating novel scenes and objects not present in the input frames, such as synthesizing a completely new environment after a turn while rendering physically plausible details like consistent reflections on the vehicle's hood. These capabilities are visually substantiated in~\figurename~\ref{fig:vis}. We compare STRIDE against related video generation models, demonstrating its advantage in preserving scene dynamics and visual fidelity in~\figurename~\ref{fig:compare}.

\begin{figure*}[t]
    \centering
    \includegraphics[width=1.0\linewidth]{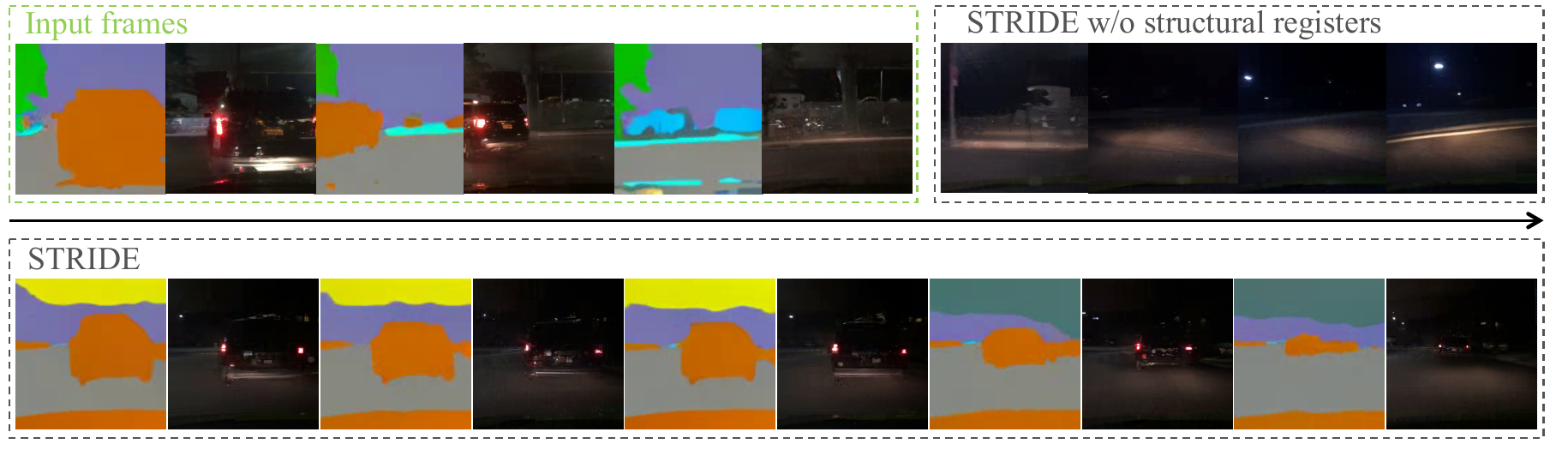}
    \caption{Qualitative ablation on structural registers. This comparison highlights the importance of the semantic guidance scheme in a challenging nighttime scenario. In such difficult conditions, the model without structural registers fails to generate a coherent new scene, losing track of the lead vehicle's presence after the turn. In contrast, STRIDE successfully maintains the vehicle's presence and structure, demonstrating that structural registers are crucial for robust temporal generation.}
    \label{fig:wosemantic}
\end{figure*}

\begin{table}[h!]
    \centering
    \caption{Quantitative ablation of STRIDE components. The table shows the impact on FID and FVD scores on the BDD100K and nuScenes datasets when the structural registers and the super-resolution module are individually removed, and compares the performance of the smaller 7B LVM variant against the full 20B STRIDE model.}
    \begin{tabular}{@{}lcccc@{}}
        \toprule
        \multirow{2}{*}{Setting} & \multicolumn{2}{c}{BDD100K} & \multicolumn{2}{c}{nuScenes} \\
        \cmidrule(r){2-3}
        \cmidrule(r){4-5}
         & FID $\downarrow$ & FVD $\downarrow$ & FID $\downarrow$ & FVD $\downarrow$ \\
        \midrule
        w/o SR module & 67.3 & 129.9 & 44.3 & 89.5 \\
        w/o registers & 35.4 & 91.6 & 30.4 & 84.2 \\
        w/ 7B LVM & 27.6 & 98.3 & 30.8 & 49.4 \\
        STRIDE & \textbf{22.4} & \textbf{80.1} & \textbf{16.7} & \textbf{40.8} \\
        \bottomrule
    \end{tabular}
    \label{tab:factors}
\end{table}

\paragraph{Stochastic and diverse future generation.} Our model demonstrates a capacity for stochastic generation, producing diverse and plausible future timelines from a single, deterministic input. To analyze this behavior, we generate $50$ distinct videos from the same $3$-frame prompt. In a specific scenario where the lead vehicle signals a right turn, the model's stochasticity is grounded in plausible outcomes. While the most probable generation is for the ego-vehicle following the lead vehicle's turn, the model also explores physically consistent alternatives, such as proceeding straight (in $34\%$ of the outcomes) or reacting to the lead vehicle temporarily aborting its turn (in $6\%$ of the outcomes). These diverse and logical outcomes, visually substantiated in~\figurename~\ref{fig:timelines}, demonstrate the model's ability to handle scene uncertainty.

\subsection{Ablation study}

We conduct an ablation study to analyze the contributions of two key components: the structural registers and the post-processing super-resolution (SR) module. They cope with orthogonal and critical problems: long-term temporal consistency and per-frame visual fidelity, respectively. We also report the comparison between the 7B LVM baseline and the full 20B STRIDE model as a capacity study that formally validates the architectural scalability of our framework.

The qualitative analysis in~\figurename~\ref{fig:wosemantic} demonstrates the benefit of structural registers. In a challenging nighttime scenario, structural registers are crucial for maintaining object permanence and generating novel scenes. This finding is substantiated by the quantitative results in~\tablename~\ref{tab:factors}. Despite the interleaved sequence being twice as long and thus a more challenging modeling task, STRIDE still achieves significantly better FID and FVD scores compared to the same model run in RGB-only mode. The performance gap is particularly pronounced on the more challenging out-of-distribution nuScenes dataset, highlighting the generalization ability imparted by structural registers. \tablename~\ref{tab:factors} also serves to quantify the impact of the SR module, validating its essential role in achieving the final visual fidelity.

%% file: sec/5_conclusion.tex
\section{Conclusion and limitation}

We develop STRIDE, a long-term video continuation framework that employs a structural foresight mechanism to fundamentally separate the prediction of high-level structure (via semantic registers) from the synthesis of high-entropy visual details. Experiments demonstrate that this novel framework is crucial for enhancing temporal consistency and mitigating generative degeneration. Furthermore, we confirm that an architecturally decoupled super-resolution module can enhance visual fidelity while significantly mitigating the computational burden of the LVM. Our work demonstrates that a purely token-based model, when guided by explicit structural priors, is an effective paradigm for generating coherent and creative front-view driving videos. Despite the efficiency gains from architectural decoupling, the primary limitation of STRIDE remains its inference speed. Future work should investigate advanced tokenization strategies to reduce training cost and inference latency. In addition, incorporating more auxiliary modalities---such as text, depth maps, trajectories, and actions---offers a promising direction for enabling broader applications in real-time, closed-loop planning and control systems.

%% file: sec/X_suppl.tex
This appendix provides supplementary material validating key components of our framework.

\section{Consistency between generated modalities}

\begin{figure}[h!]
    \centering
    \includegraphics[width=0.6\linewidth]{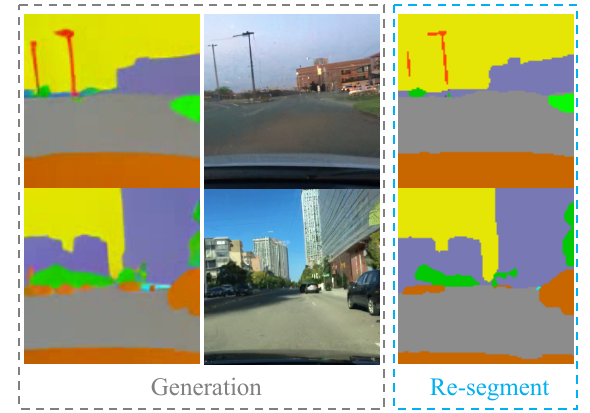}
    \caption{\textbf{Validating semantic foresight through a closed-loop re-segmentation experiment.} The left column (\textit{Generation}) shows the semantic map generated by STRIDE, while the right column (\textit{Re-segment}) shows the results of feeding the generated RGB frames back into the same semantic segmentation model (UniFormer). The strong correspondence validates that our model maintains strict consistency between the predicted semantic registers and the final visual appearance.}
    \label{fig:consistency}
\end{figure}

To rigorously verify the alignment between the generated structural registers and visual appearance, we conducted a closed-loop re-segmentation experiment, as illustrated in~\figurename~\ref{fig:consistency}. In this experiment, the RGB frames generated by STRIDE are fed back into the same pre-trained UniFormer model originally employed for semantic segmentation. We observe that the resulting re-segmented outputs align closely with the semantic maps generated by STRIDE. This confirms that STRIDE accurately translates structural registers into visual pixels without hallucinating conflicting geometries. This qualitative finding is further corroborated by a quantitative evaluation on the nuScenes validation set, where our approach achieves a $77.4\%$ mean IoU, evidencing the robust cross-modal consistency maintained by STRIDE.

\section{Ablation of the super-resolution module}

\begin{figure}[h!]
    \centering
    \includegraphics[width=\linewidth]{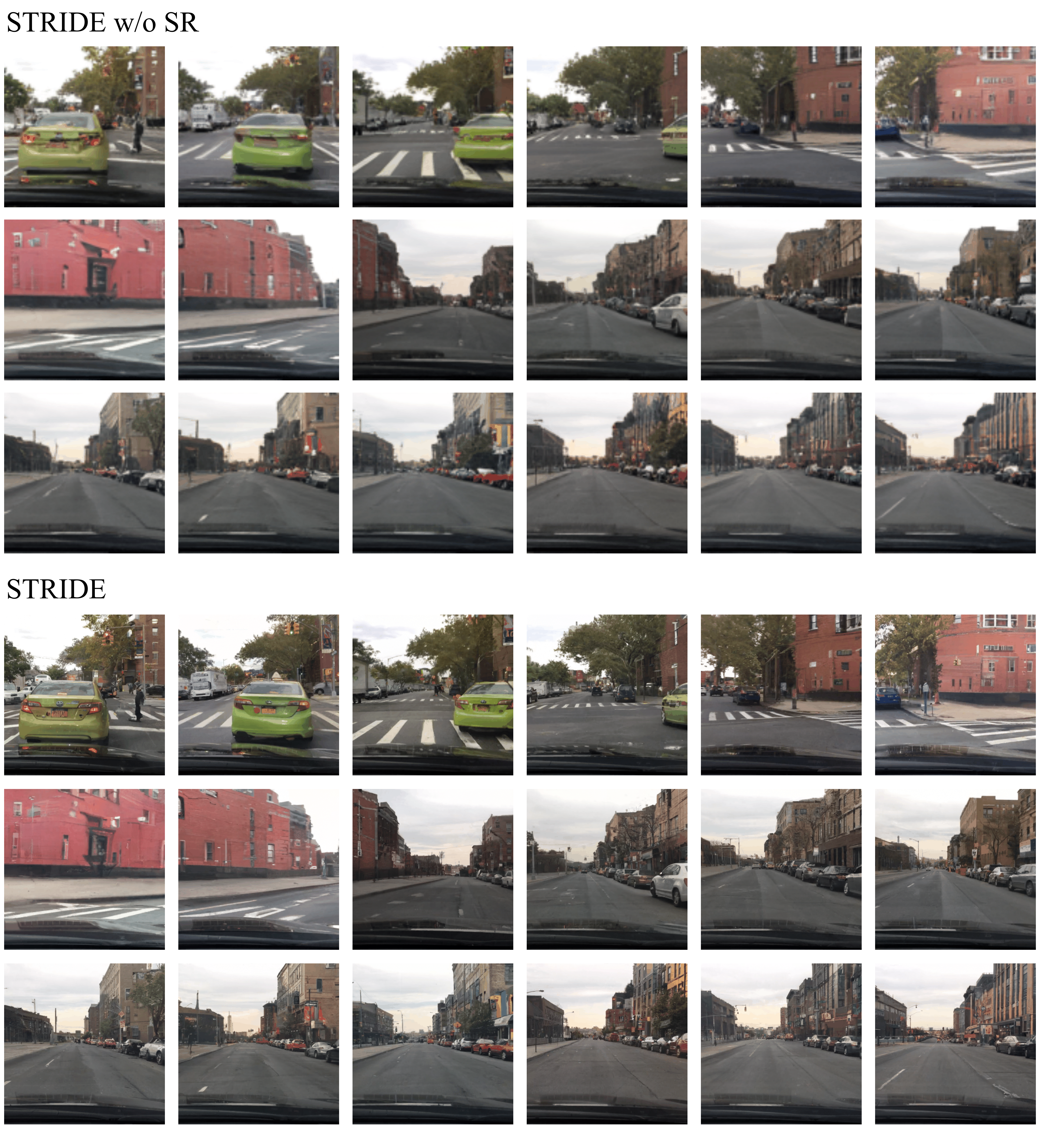}
    \caption{\textbf{Qualitative ablation of the post-processing super-resolution strategy.} STRIDE decouples structural generation from detail rendering. The top rows (\textit{STRIDE w/o SR}) show the raw output of the LVM upscaled via bilinear interpolation, revealing that the LVM correctly captures the scene structure but lacks pixel-level fidelity. The bottom rows (\textit{STRIDE}) confirm that our flow matching-based SR module effectively generates plausible high-frequency details, significantly enhancing perceptual quality.}
    \label{fig:AR+SR}
\end{figure}

The incorporation of a post-processing super-resolution (SR) module is a strategic design choice that alleviates the large vision model (LVM) from the burden of rendering fine-grained details, thereby allowing it to dedicate its capacity to generating structurally coherent low-resolution videos. The qualitative ablation presented in \figurename~\ref{fig:AR+SR} validates the efficacy of this approach. While the baseline frames upscaled via bilinear interpolation maintain structural integrity, they inherently lack high-frequency details. In contrast, frames processed by our flow matching-based SR module exhibit significantly enhanced visual quality, characterized by sharper edges and enriched textures. This comparison confirms the module's indispensable role in achieving high-fidelity results.

\section{Implementation details}
\label{app:impl}

This section documents implementation details of the tokenization, data preparation, inference, super-resolution, and analysis procedures that complement the main text.

\subsection{Tokenization and sequence format}
\label{app:tokenization}

\paragraph{Visual tokenizer.} We use the publicly released Open-MAGVIT2 tokenizer trained on ImageNet at $128 \times 128$ resolution~\citep{luo2024open}. Its encoder has three down-sampling stages, i.e., a spatial compression factor of $8$, and quantizes each latent position into an $18$-bit lookup-free code ($2^{18} = 262{,}144$ entries). The encoder and decoder together have $63.6$M parameters, and all tokenizer weights are kept frozen. Following the token factorization of MAGVIT-v2~\citep{yu2023language}, each $18$-bit code is split into two $9$-bit sub-tokens. For a $112 \times 112$ frame, the $14 \times 14$ latent grid therefore yields $392$ tokens: the $196$ first sub-tokens in raster order, followed by the $196$ second sub-tokens in raster order. The two sub-token types occupy disjoint index ranges, so the LVM operates on an image-token vocabulary of $1{,}024$ entries. Within each timestep, the $392$ tokens of the semantic map precede the $392$ tokens of the RGB frame. \Cref{tab:token_budget} summarizes the resulting token budget.

\begin{table}[h]
    \centering
    \caption{Token budget of the LVM under the context length of $16{,}384$ tokens. The last row shows the budget that direct modeling at the super-resolved output size would require with the same tokenizer.}
    \label{tab:token_budget}
    \begin{tabular}{@{}lcccccc@{}}
        \toprule
        \multirow{2}{*}{Frame size} & \multirow{2}{*}{Latent grid} & \multirow{2}{*}{Tokens / image} & \multicolumn{2}{c}{Tokens / timestep} & \multicolumn{2}{c}{Max.\ timesteps} \\
        \cmidrule(lr){4-5} \cmidrule(l){6-7}
         & & & RGB-only & Interleaved & RGB-only & Interleaved \\
        \midrule
        $112 \times 112$ & $14 \times 14$ & $392$ & $392$ & $784$ & $41$ & $20$ \\
        $448 \times 448$ & $56 \times 56$ & $6{,}272$ & $6{,}272$ & $12{,}544$ & $2$ & $1$ \\
        \bottomrule
    \end{tabular}
\end{table}

\paragraph{Semantic maps.} Semantic maps are produced by the UniFormer-S model with an UperNet head trained on ADE20K ($150$ classes)~\citep{li2023uniformer}, using the checkpoint distributed with the ControlNet annotators~\citep{zhang2023adding}. Segmentation is run on full-resolution frames, and each label map is rendered as a color image with the ADE20K palette before it is cropped, resized, and tokenized in exactly the same way as the corresponding RGB frame. At inference time, UniFormer is applied only to the three prompt frames: all subsequent semantic maps are generated by the LVM, and no ground-truth semantic information of future frames is used in any comparison.

\subsection{Data preparation}
\label{app:data}

BDD100K videos are decoded at their native frame rate (predominantly $30$ FPS), and every tenth frame is kept, which yields approximately $3$ Hz. Each frame is center-cropped to a square whose side equals the shorter image side, resized to $112 \times 112$ pixels, and normalized to $[-1, 1]$ before tokenization. With roughly $110$ frames per training video, the RGB and semantic streams each contribute about $3$ billion tokens, which constitutes the corpus of approximately $6$ billion tokens reported in the main text. Videos of the nuScenes validation set are cropped and resized in the same way.

\subsection{Inference procedure}
\label{app:inference}

\Cref{alg:rollout} summarizes auto-regressive continuation. The tokens of the three prompt frames and their semantic maps form the initial context. The LVM then generates one timestep at a time, first the semantic tokens and then the RGB tokens, and every generated timestep is appended to the context before the next one is generated. The generated RGB tokens are decoded by the frozen tokenizer decoder and passed to the super-resolution module.

Minute-level rollouts, such as the one in \figurename~\ref{fig:auto}, are generated in consecutive windows. Each window is prompted with three timesteps and generates the next $15$; the following window starts from a fresh context whose prompt consists of the last three generated timesteps, including their semantic tokens. Three windows thus produce $3 + 3 \times 15 = 48$ timesteps. With the 7B model, decoding runs at about $60$ tokens per second, i.e., about $6.5$ seconds for the $392$ tokens of one image and about twice as long for an interleaved timestep.

\begin{algorithm}[h]
\caption{Interleaved auto-regressive continuation with structural registers.}
\label{alg:rollout}
\begin{algorithmic}[1]
\Require Prompt frames $I_1, I_2, I_3$; segmentation model $\mathrm{Seg}$; tokenizer $\mathcal{T}$ with decoder $\mathcal{D}$; LVM $p_\theta$; super-resolution module $\mathrm{SR}$; number of windows $W$
\State $C \gets \big[\mathcal{T}(\mathrm{Seg}(I_1)), \mathcal{T}(I_1), \mathcal{T}(\mathrm{Seg}(I_2)), \mathcal{T}(I_2), \mathcal{T}(\mathrm{Seg}(I_3)), \mathcal{T}(I_3)\big]$ \Comment{semantic maps only for the prompt}
\For{$w = 1, \dots, W$}
    \For{$k = 1, \dots, 15$}
        \State $\hat{Z}^S \sim p_\theta(\cdot \mid C)$ \Comment{$392$ semantic tokens, sampled token by token}
        \State $\hat{Z}^I \sim p_\theta(\cdot \mid C, \hat{Z}^S)$ \Comment{$392$ RGB tokens}
        \State $C \gets C \oplus [\hat{Z}^S, \hat{Z}^I]$
        \State \textbf{emit} $\mathrm{SR}\big(\mathcal{D}(\hat{Z}^I)\big)$
    \EndFor
    \State $C \gets$ the last three generated timesteps (semantic and RGB tokens) \Comment{context for the next window}
\EndFor
\end{algorithmic}
\end{algorithm}

\subsection{Flow matching-based super-resolution}
\label{app:sr}

The velocity network $v_\theta$ is an EfficientUNet with $305.8$M parameters, using channel multipliers $(1, 2, 4, 8)$ and three residual blocks per resolution level. Its input is the channel-wise concatenation of the current state $x_t$ and the context $c = x_l$, where $x_l$ is the bilinear $4\times$ upsampling of a decoded $112 \times 112$ frame, and its output is a three-channel velocity. The initial state $z$ is obtained by applying the forward process of a $1000$-step cosine noise schedule to $x_l$ at step $400$, i.e., $z = \sqrt{\bar{\alpha}}\, x_l + \sqrt{1-\bar{\alpha}}\, \epsilon$ with $\bar{\alpha} \approx 0.65$ and $\epsilon \sim \mathcal{N}(0, \mathbf{I})$. The flow follows the linear path $x_t = t\, r + (1-t)\, z$ with target velocity $u_t = r - z$, and $t$ is sampled uniformly from $[0, 1]$ during training. At inference, the ODE is integrated from $t = 0$ to $t = 1$ with a fixed-grid fourth-order Runge--Kutta scheme on $20$ uniformly spaced time points, and the resulting residual is added to $x_l$ (\Cref{alg:sr}). Given the initial state, the integration is deterministic. The module processes each frame independently, including the prompt frames.

\begin{algorithm}[h]
\caption{Flow matching-based super-resolution of one frame.}
\label{alg:sr}
\begin{algorithmic}[1]
\Require Decoded frame $y \in \mathbb{R}^{112 \times 112 \times 3}$; velocity network $v_\theta$; number of time points $N = 20$
\State $x_l \gets \mathrm{BilinearUpsample}(y, 448 \times 448)$
\State $x \gets \sqrt{\bar{\alpha}}\, x_l + \sqrt{1-\bar{\alpha}}\, \epsilon$, \quad $\epsilon \sim \mathcal{N}(0, \mathbf{I})$ \Comment{initial state $z$}
\State $t_n \gets n / (N-1)$ for $n = 0, \dots, N-1$
\For{$n = 0, \dots, N-2$}
    \State $x \gets \mathrm{RK4Step}\big((x', t') \mapsto v_\theta(x', x_l, t'),\; x,\; t_n,\; t_{n+1}\big)$
\EndFor
\State \Return $\hat{x}_h \gets x_l + x$ \Comment{$x$ approximates the residual $r$}
\end{algorithmic}
\end{algorithm}

\subsection{Analysis metrics}
\label{app:metrics}

\paragraph{Motion intensity.} For \figurename~\ref{fig:probing}, optical flow between consecutive frames is estimated with RAFT~\citep{teed2020raft}, using the checkpoint trained on FlyingChairs and FlyingThings3D and $20$ refinement iterations, after resizing the frames to $224 \times 224$ pixels. For each pair of consecutive generated frames, we average the Euclidean norm of the flow vectors over all pixels and then average over the $150$ clips. The ground-truth reference is the mean flow magnitude of the real clips, which is why it is drawn as a horizontal line.

\paragraph{Codebook usage.} For \tablename~\ref{tab:codebook}, we count how often each of the $2^{18}$ codes occurs in the tokenized RGB and semantic streams, sort the codes by frequency, and report the smallest fraction of the codebook whose occurrences account for the given proportion of all tokens.

\paragraph{Stochastic outcomes.} The outcome statistics accompanying \figurename~\ref{fig:timelines} are obtained by manually categorizing $50$ rollouts generated from the same prompt according to the maneuvers of the ego vehicle and the lead vehicle.

\section{Broader impact statement}
\label{app:impact}

Generative driving video models can support simulation, data augmentation, and the analysis of rare traffic scenarios. At the same time, realistic synthetic dash-camera footage could be misused to fabricate evidence of traffic events. Synthetic data may also contain plausible-looking but incorrect behaviors, and relying on it to validate safety-critical systems without real-world verification could create a false sense of reliability. Finally, the training data was collected in a limited set of geographic regions, so the learned driving behaviors and scene statistics may not transfer to other regions. We recommend clearly disclosing synthetic content and validating downstream autonomous-driving components on real-world data.